\documentclass[letterpaper, 10 pt, conference]{ieeeconf}  

\IEEEoverridecommandlockouts                              

\usepackage{graphicx}
\usepackage{amsfonts}
\usepackage{hyperref}
\usepackage{diagbox}
\usepackage{stfloats}
\usepackage{caption}
\DeclareCaptionFont{mysize}{\fontsize{8}{9.6}\selectfont}
\usepackage[ruled,linesnumbered]{algorithm2e}

\usepackage{amsmath} 
\usepackage{amssymb}  
\usepackage{listings}
\usepackage{booktabs}
\usepackage{makecell}
\usepackage{etoolbox}
\usepackage{color}
\definecolor{blue}{RGB}{44,105,141}
\definecolor{orange}{RGB}{252,148,45}
\usepackage[dvipsnames]{xcolor}

\makeatletter
\patchcmd{\@makecaption}
  {\scshape}
  {}
  {}
  {}
\makeatother
\title{\LARGE \bf

Hierarchical Reinforcement Learning Based on Planning Operators*
}

\author{Jing Zhang, Emmanuel Dean, and Karinne Ramirez-Amaro
\thanks{*This work is supported by Chalmers AI Research Centre (CHAIR).}
\thanks{Jing Zhang, Emmanuel Dean and Karinne Ramirez-Amaro are with the Faculty of Electrical Engineering, 
        Chalmers University of Technology, SE-412 96 Gothenburg, Sweden.
        {\tt\small \{jingzhan, karinne\}@chalmers.se}}%
}

\begin{document}

\maketitle
\thispagestyle{empty}

\pagestyle{empty}

\begin{abstract}
Learning long-horizon manipulation tasks such as stacking, presents a longstanding challenge in the field of robotic manipulation, particularly when using Reinforcement Learning (RL) methods. RL algorithms focus on learning a policy for executing the entire task instead of learning the correct sequence of actions required to achieve complex goals. While RL aims to find a sequence of actions that maximises the total reward of the task, the main challenge arises when there are infinite possibilities of chaining these actions (e.g. reach, grasp, etc.) to achieve the same task (stacking). In these cases, RL methods may struggle to find the optimal policy. This paper introduces a novel framework that integrates the operator concepts from the symbolic planning domain with hierarchical RL methods. We propose to change the way complex tasks are trained by learning independent policies of the actions defined by high-level operators instead of learning a policy for the complete complex task. Our contribution integrates planning operators (e.g. preconditions and effects) as part of the hierarchical RL algorithm based on the Scheduled Auxiliary Control (SAC-X) method. We developed a dual-purpose high-level operator, which can be used both in holistic planning and as independent, reusable policies. Our approach offers a flexible solution for long-horizon tasks, e.g., stacking and inserting a cube. The experimental results show that our proposed method achieved an average success rate of $97.2\%$ for learning and executing the whole \texttt{stack}. Furthermore, we obtain a high success rate when learning independent policies, e.g. reach (98.9\%), lift (99.7\%), move (97.4\%), etc. The training time is also reduced by $68\%$ when using our proposed approach. 
\end{abstract}

\section{Introduction}
Long-horizon tasks, such as stacking blocks or assembling products, present a particularly daunting challenge for autonomous systems. In deep reinforcement learning (DRL), agents must learn the intricate dependencies between actions and determine a viable sequence for execution. Random exploration is hardly sufficient for this, making data collection for future learning ineffective due to the decreased likelihood of encountering meaningful rewards. While methods like behavioral cloning (BC) \cite{b1}, \cite{b2}, offer an intuitive approach by mimicking human actions, they come with their own set of limitations, e,g. a set of several human demonstrations needed. Training human experts to provide informative demonstrations is not only labor-intensive but also lacks generalizability across various tasks since it is not possible to demonstrate all possible exceptions.

Learning a sequence of actions is challenging and the symbolic planning methods address this problem. Planners serve as a structured and scalable framework for tackling complex, long-horizon operations. While the foundational knowledge for symbolic structures is often provided by humans, this approach tends to be more interpretable. It contains the definition of preconditions, effects, and pathways from the initial to the goal state. However, this method is not without its drawbacks. A key limitation lies in the necessity for an exhaustive and accurate delineation of the planning domain. Manual construction of such comprehensive domains is often impractical due to the considerable time and resources required, and in some cases, it may even be unfeasible, as pointed out in \cite{jimenez2012review}. Therefore, the challenge of finding an efficient and universally applicable solution remains.
\begin{figure}[t]
  \centering
  \includegraphics[width=0.45\textwidth]{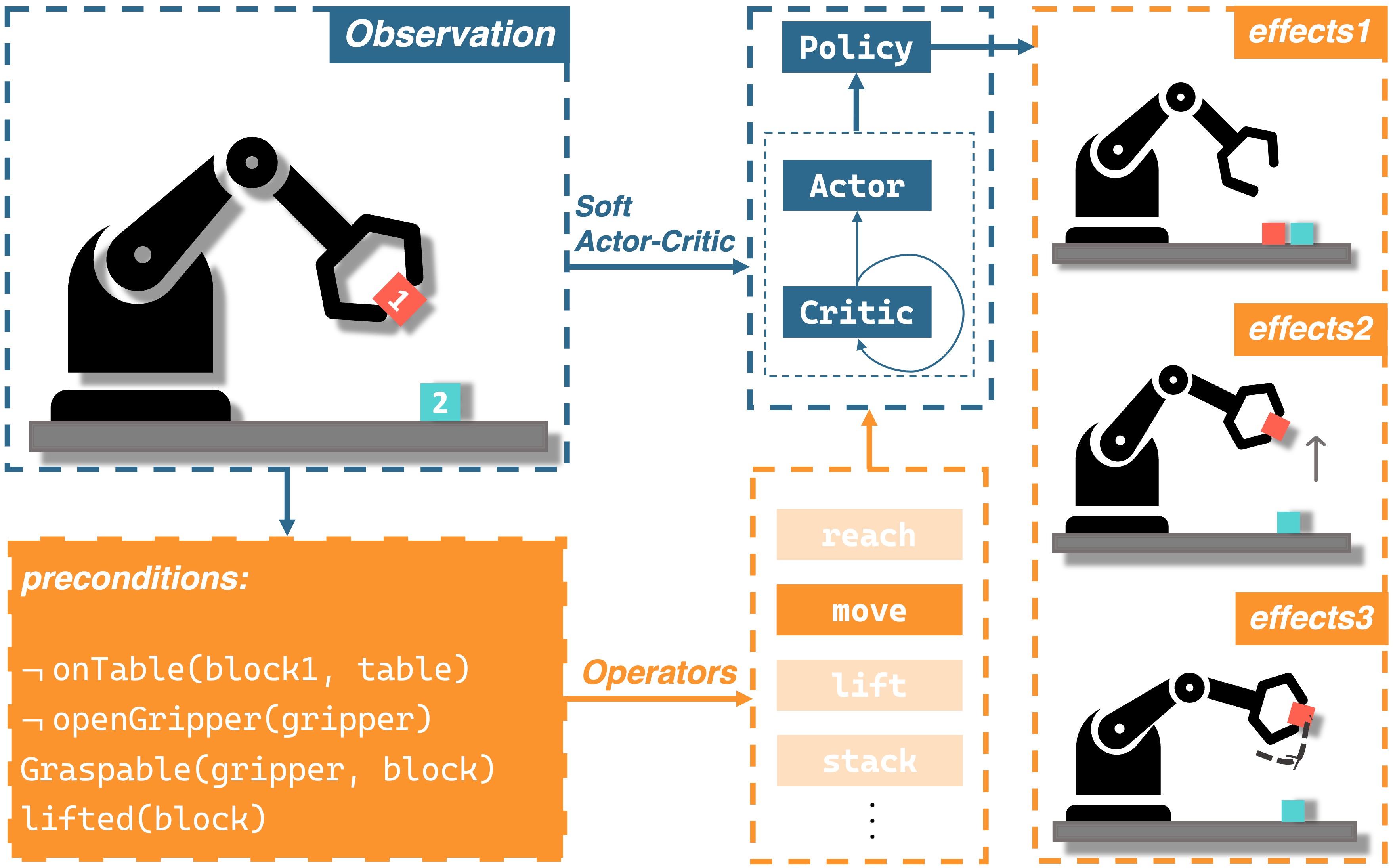}
  \caption{Overview of our approach of combining RL and planning operators.}
  \label{fig:main}
\end{figure}

In this paper, we will address some of the limitations of both RL and planners in a holistic approach (see Fig. \ref{fig:main}). Particularly, we will use the planning operators concepts from symbolic planning to build the hierarchy of actions used in the hierarchical RL method Schedule Auxiliary Control (SAC-X) \cite{riedmiller2018learning}. First, we extract the environmental state and knowledge of its dynamics encoded as preconditions and effects. These state-based observations are used to define the plan operators which are used to generate the sequence of actions an agent should execute to achieve the desired goal. This information is then used by the RL agent to learn low-level policies to execute the needed actions independently. Thus these operators could be dynamically selected based on the object's state. Operators also can be reused when the object's number is enlarged or the agent is assigned to a new task. In this work, we define operators that are independent, meaning that each operator will learn a low-level control policy. The goal is to learn policies that can be used even without a planning process. Low-level policies are responsible for executing specific manipulator translational movements in a physical context. Using hierarchical reinforcement learning (HRL) enables the robot to learn how to plan the right execution order for long-horizon tasks via high-level operators and apply it through low-level manipulator policy. We will validate our method in a stacking and inserting cube scenario. To summarize, our main contribution is to propose a novel learning method based on hierarchical RL and symbolic planning to learn independent policies and automatically construct the right action sequence using preconditions and effects for achieving the long-horizon manipulation goal. 

\section{Related Works}\label{related}
Our research draws inspiration from areas in reinforcement learning and automated planning. In the field of task-level planning, classic methods such as symbolic planning \cite{fox2003pddl2} and algorithms such as Probabilistic Random Maps (PRM) or Rapid Exploration Random Trees (RRT) have been prominent for low-level task solver. These methods are often domain-specific and rely heavily on expertise and hard-coded rules. M. Diehl et al. \cite{diehl2021automated} investigated the possibility of automatically generating robotic planning domains from human observations where an additional planning domain definition language (PDDL) solver is needed to handle domain-based problems. However, this work does not deal with the problem of learning low-level skill execution which is provided manually. R. Takano et al. \cite{RSL} presents an approach for skill learning in Task and Motion Planning (TAMP) by incorporating Bayesian Neural Networks (BNNs) based Level Set Estimation (LSE) \cite{ha2021high}. Influenced by Guided Policy Search (GPS) \cite{montgomery2016guided}, their framework adopts a hierarchical structure where RL is used for data collection and then employs supervised learning for fine-tuning the policy. One of the lingering challenges in their approach is how to efficiently generate training data. 

Building on the theme of hierarchical structures in RL, M. Riedmiller et al. \cite{riedmiller2018learning} proposed a deep reinforcement learning (DRL) paradigm - Scheduled Auxiliary Control (SAC-X) that enables learning of complex behaviors from scratch in the presence of multiple spare rewards. This method contains a hierarchical agent to organize high-level scheduler and low-level intentions. However, the computational cost remains a challenge, as evidenced by the total 5000 episodes $\times$ 36 actors $\times$ 360 time steps = 64.8M time steps required for learning a stacking task. One reason that causes the RL algorithm to struggle in learning long-horizon manipulation tasks is that state transition in MDP has randomness, resulting in difficulties for RL agent to learn the right action sequence that could achieve the goal. Therefore, we are proposing to use symbolic planning to tackle this issue.

Symbolic planning could provide guidance and has proven to be a feasible method when combined with RL \cite{illanes2020symbolic}, while these existing approaches have been effective in generating holistic plans for such tasks \cite{SDRL}, \cite{guan2022leveraging}, they often encapsulate operators within the planning framework, making it infeasible to utilize these operators in isolation for other tasks or scenarios. In contrast, our work offers a level of modularity and flexibility. Specifically, our method not only integrates high-level operators derived from symbolic methods but also enables the generated high-level operators to function both within the scope of a specific plan and as separate, reusable policies. This unique feature addresses the long-standing challenge in RL of learning the action sequences for long-horizon tasks while offering a greater range of applicability.
\section{Method}\label{method}

\subsection{Preliminaries}\label{preliminary}
\subsubsection{Reinforcement learning}
\begin{figure}[h]
  \centering
  \includegraphics[width=0.45\textwidth]{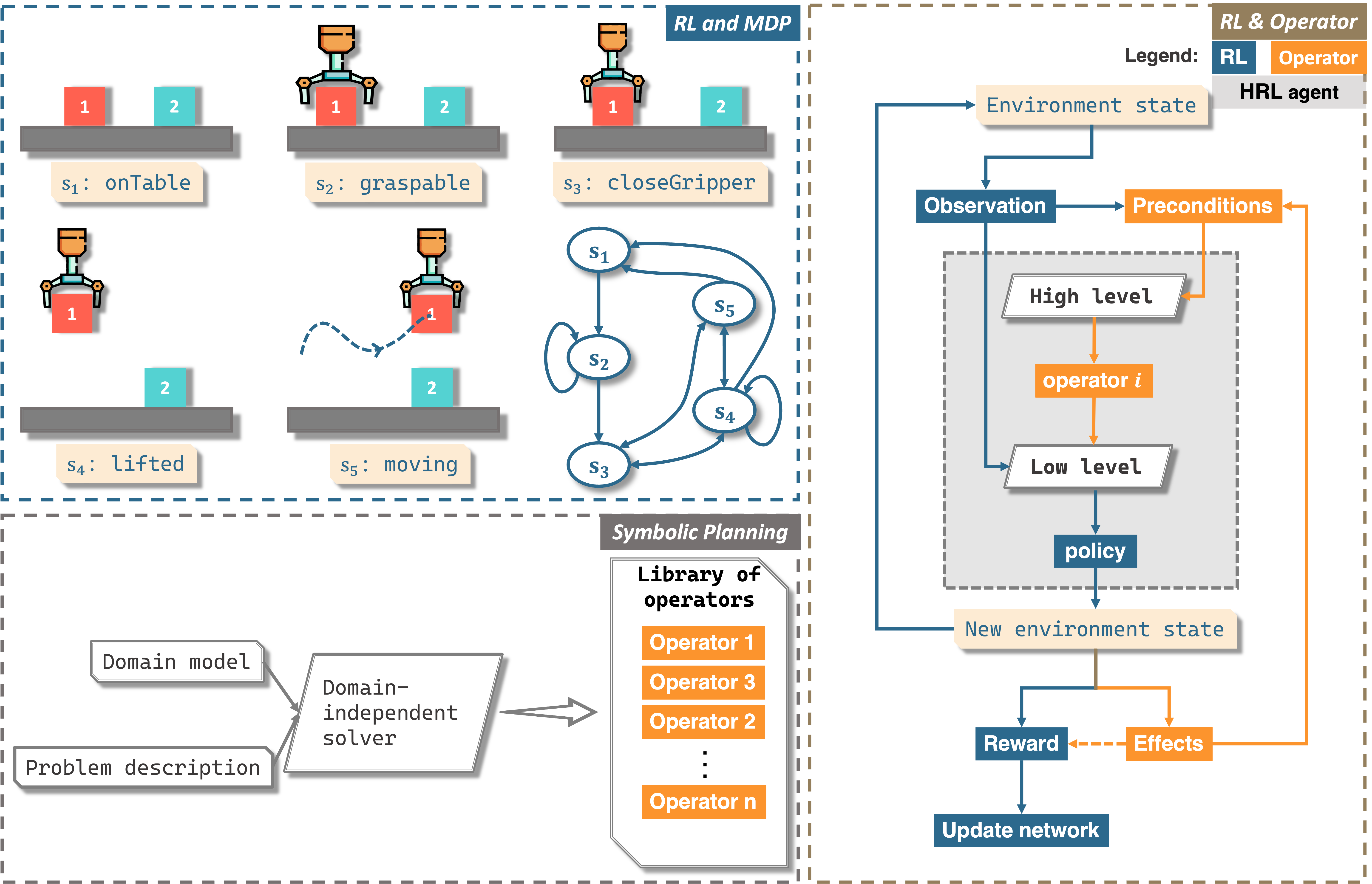}
  \caption{Figures on the left show the traditional RL and a symbolic planning framework while the figure on the right shows the framework we proposed in this paper that combines RL and Operators. Here $s$ is used to describe the state related to blocks or gripper.} 
  \label{fig:des}
\end{figure}

A Markov decision process (MDP) is defined as $\mathcal{M}=\{\mathcal{S}, \mathcal{A}, \mathcal{P}, r, \gamma\}$, here $\mathcal{S}$ and $\mathcal{A}$ are the sets of state and actions respectively, $\mathcal{P}$ is the state-transition kernel $\mathcal{P}:\mathcal{S}\times\mathcal{A}\times\mathcal{S}\rightarrow[0,1]$ and  $r(s,a):\mathcal{S}\times\mathcal{A}\rightarrow \mathcal{R}$ a possibly reward function. $\gamma$ is the discount factor that ranges between $0$ and $1$. The solution of the MDP is a policy $\pi: \mathcal{S}\rightarrow\mathcal{A}$ mapping a state to an action, actions are sampled from a policy $\pi(a|s)$. RL aims at maximizing the value function $V^{\pi} = \mathbb{E}_{\pi}[\sum_{t=0}^{t_{f}}\gamma^t\mathcal{R}(s_t,a_t)]\ |\ a_t\sim \pi(\cdot|s_t), s_{t+1}\sim \mathcal{P}(\cdot|s_t,a_t)$, where $t$ refers to time step and $t_{f}$ is the time that the goal is achieved or when it reaches the end of assigned time period for one trajectory. Fig. \ref{fig:des} \textit{RL and MDP} part shows a simple 5-state MDP. In long-horizon tasks, such as the manipulation of complex objects in dynamic environments, RL algorithms face challenges in identifying the optimal sequence of actions to reach the goal. One of the primary reasons is the probabilistic nature of state transitions. Because of the randomness, the same action can lead to different subsequent states, i.e., in Fig. \ref{fig:des}, starting from $s_4:lifted$, the next state could be $s_1,s_3,s_5$ or even $s_4$ itself. This unpredictability complicates the learning process, as the RL agent must explore various pathways and sequences without any guidance to arrive at an effective policy for action selection, especially for complex tasks. Thus, we propose to use the expressiveness of high-level methods such as symbolic planning to bootstrap the learning of the next action.
\vspace{-0.3cm}
\subsubsection{Symbolic planning} focuses on generating a sequence of high-level operators\footnote{Operators are also known as actions, here we use the term operators to discriminate from the action concepts used in RL.} to achieve a specific goal. Symbolic planning employs symbolic language to define operators through preconditions and effects, then, it reasons about these symbols to infer a sequence of actions by using a planner algorithm. Such planners require additional information about the initial and goal states to obtain a plan. A planning task is typically divided into two main components: the domain and the problem. The \textit{problem} contains the initial condition and goal that needs to be reached while \textit{domain} offers a comprehensive list of all feasible operators that can be taken to transition from the initial state to the desired goal state. In this work, we are only using the definition of the \textit{domain} to obtain the list of learned operators for tasks. One advantage of using the \textit{domain} is that it can be applied to multiple problems as long as they operate within the same set of state variables and actions.

\begin{figure}[thpb]
  \centering
  \includegraphics[width=0.40\textwidth]{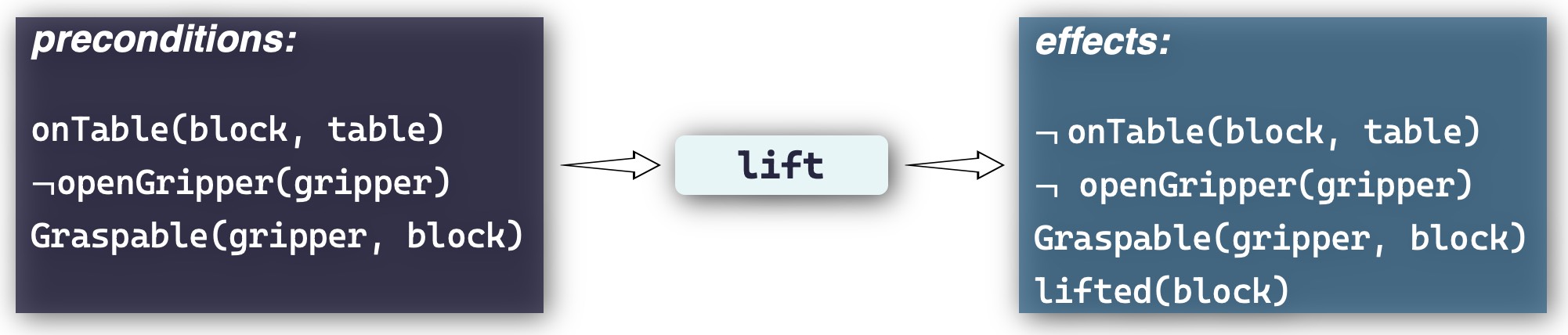}
  \caption{Operator structure which is composed of preconditions and effects.}
  \label{fig:operator}
\end{figure}

In symbolic planning, operators are templated action schematics that, when applicable, modify the state of the environment in a predetermined manner. A unique name identifies each planning operator and consists of three key elements: i) \textit{Arguments}: a collection of objects that serve as parameters for the operator, specifying its context; ii) \textit{Preconditions}: these are logical conditions or criteria that must hold \textit{True} in the current environment state for the operator to be applicable; and iii) \textit{Effects}: these are the outcomes that occur in the environment state after the execution of the operator. Like preconditions, effects are expressed in a formalized way to predict how the environment will transition as a result of applying the operator.
Both preconditions and effects use binary state variables, often referred to as predicates, which help in defining the conditions and consequences in a more structured manner. For example the operator (\verb|lift|) is shown in Fig. \ref{fig:operator}.

For the definition of \textit{domain}, we followed the convention of the Plan Domain Definition Language (PDDL) \cite{fox2003pddl2} to formulate a library of operators. A \textit{domain} is defined as $\sum=(\mathcal{S}, \mathcal{O}, \Gamma, cost)$. $\mathcal{S}$ is the set of state variables which can be \textit{True} or \textit{False}, $\mathcal{O}$ is the set of operators, $\Gamma:\mathcal{S}\times\mathcal{O}$ is the state-transition function and $cost: \mathcal{S}\times\mathcal{O}\rightarrow[0, \infty]$ is a cost function. State variables describe relations between environment objects.

As shown in Fig. \ref{fig:des}, we are using the definition of \textit{domain} from \textit{PDDL} to define a set of operators. The symbolic description of operators allows interpretability and transferability at a high level, however, the obtained operators do not contain information on low-level parametrization needed for the robot execution. For instance, consider a stack scenario where an operator is defined as 'reach' but this operator does not contain the specifics of how the manipulator should reach the target—such as the required position, velocity, or orientation. This limitation highlights one of the challenges of applying symbolic planning definitions in scenarios that require low-level control details. Therefore, in this paper, we are proposing a solution.

\subsection{Hierachical Reinforcement learning based on operators}\label{hrl}
In this work, we adjusted the SAC-X algorithm \cite{riedmiller2018learning} to learn the agent's hierarchy by integrating the concepts of planning operators as part of the MDP definitions. The operators are denoted as $\mathcal{O}=\{\mathcal{O}_1,\cdots, \mathcal{O}_K\}$, where $\mathcal{O}$ refers to a set of $K$ operators. These enhanced MDPs share state, action space, transition function, and hyperparameters, but diverge in their respective reward function. Our modified SAC-X method learns low-level control policies by maximizing the task return $J(\pi_{\mathcal{O}})$ (see eq. \ref{eq:J}) and it uses the planning operators to define the high-level scheduler policy. The scheduler has the important role of maximizing the expected return (see eq. \ref{eq:E}) of the obtained plan (hierarchy) by selecting the sequence of low-level control policies at run-time, instead of manually provided as originally proposed in \cite{riedmiller2018learning}. 
\vspace{-0.2cm}
\begin{equation} \label{eq:E}
    \mathbb E_{\pi(a|s,\mathcal{O})}\left[\mathcal{R}_{\mathcal{O}}(\pi_{t:t_{f}})\right]=\mathbb E_{\pi_(a|s,\mathcal{O})}\left[\sum_{t=0}^{t_{f}}\gamma^t\mathcal{R}(s_t,a_t)\right]
\end{equation}
To learn the low-level policies, we used the Soft Actor-Critic (SAC) method \cite{haarnoja2018soft}, which is a model-free off-policy algorithm that maximizes the entropy ($\mathcal{H}$) of the policy:
\vspace{-0.15cm}
\begin{equation}
    \mathcal{H}(X)=\mathbb E_{(x\sim p)}\left[ -\log p(x)\right]
\end{equation}
where $X$ indicates a stochastic variable with $p$ to represent the variable probability density function.
In RL, $\mathcal{H}(\pi(\cdot |s))$ measures the policy $\pi$ randomness under state $s$, and the regularized entropy in SAC is the second term in objective function eq. \ref{eq:J}
\vspace{-0.15cm}
\begin{equation} \label{eq:J}
J(\pi_{\mathcal{O}})=\mathbb{E}_{\pi_{\mathcal{O}}}\left[\sum_{t=0}^\infty \gamma^t(R_{\mathcal{O}}(s_t, a_t) + \alpha \mathcal{H}(\pi_{\mathcal{O}}(\cdot |s_t))) \right]
\end{equation}
where $\alpha$ is a regularization coefficient that determines the importance of entropy, also called temperature. The entropy regularization increases the degree of exploration of the RL. A larger value of $\alpha$ enhances the exploratory behavior early in the learning process, which can help prevent the policy from converging prematurely to local optimum solutions. The soft Q-function is defined as:
\vspace{-0.15cm}
\begin{equation}
\begin{aligned}
Q_{\mathcal{O}}(s_t,& a_t)=R_{\mathcal{O}}(s_t,a_t)\\
+\mathbb{E}_{\pi_{\mathcal{O}}}&\left[\sum_{t=0}^\infty \gamma^t(R_{\mathcal{O}}(s_{t+1}, a_{t+1}) + \alpha \mathcal{H}(\pi_{\mathcal{O}}(\cdot |s_{t+1}))) \right]
\end{aligned}
\end{equation}
SAC models two action-value functions $Q$ and a policy function $\pi$. Based on the idea of Double DQN \cite{van2016deep}, a network with a smaller $Q$ value is selected to alleviate the problem of overestimation of the $Q$ value. The loss function is defined in eq. (5) and it aims to maximize the joint policy objective as done in \cite{ablett2023learning}: 
\vspace{-0.15cm}
\begin{equation}
\begin{aligned}
    \mathcal{L}(\pi)\!=\!\sum_{\mathcal{O}\in \mathcal{O}_{all}}\!\mathbb{E}_{s\sim\mathcal{B},a\sim\pi_\mathcal{O}(\cdot|s)}\left[Q_\mathcal{O}(s,\!a)\!-\!\alpha\! \log \!\pi_{\mathcal{O}}(a|s)\right]
\end{aligned}
\end{equation}
where $\pi$ refers to the set of policies $\{\pi_{\mathcal{O}_1}, \pi_{\mathcal{O}_2}, \cdots, \pi_{\mathcal{O}_K}\}$, $\mathcal{B}$ is the buffer that contains all of the transitions from the interaction with the environment, starting from an initial state $s_0\sim p_0$, following the policy $\pi(\cdot|s)$. By training each policy on states sampled according to the state visitation distribution of each possible task, these policies can successfully complete their respective tasks, regardless of the state in which the system was left by the policy for the previous task \cite{riedmiller2018learning}, so safely combine these learned policies is possible.

For policy evaluation, soft Q functions aim to minimize the joint Bellman residual:
\vspace{-0.15cm}
\begin{equation}
    \begin{aligned}
        \mathcal{L}(Q)&=\sum_{\mathcal{O}\in \mathcal{O}_{all}}\mathbb E_{(s,a,s')\sim\mathcal{B},a'\sim \pi_{\mathcal{O}}(\cdot|s')}\left[(Q_{\mathcal{O}}(s,a)-\delta_{\mathcal{O}})^2\right]\\
        \delta_{\mathcal{O}}&=\mathcal{R}_{\mathcal{O}}(s,a)+\gamma(Q_{\mathcal{O}}(s',a')-\alpha \log \pi_{\mathcal{O}}(a'|s'))
    \end{aligned}
\end{equation}
\vspace{-0.7cm}
\subsection{Combining RL and Symbolic planning}\label{combine}
For complex tasks, often the RL algorithm struggles to learn the right action sequence that could achieve the goal \cite{yang2021hierarchical}. It has been demonstrated that RL can learn policies for simpler tasks \cite{li2023understanding}. We hypothesise that complex tasks should be split into simple and independent actions. Then, a different approach should be used to chain the actions to achieve the desired task.  In this paper, we propose a method that tests the above hypothesis by combining RL with symbolic planning for a hierarchical RL approach as shown in Fig. \ref{fig:des}, the \textit{RL \& Operator} part. In our proposed method, each operator's effects would serve as preconditions for the subsequent operator, allowing the high-level agent to choose operators from preconditions. We define the new planning MDPs problem as $\mathcal{M}=\{\mathcal{S}, \mathcal{O}, \mathcal{P}, a, r, \gamma\}$, where $\mathcal{O}=\{\mathcal{O}_1, \cdots,\mathcal{O}_i,\cdots,\mathcal{O}_K\}$ is the set of primitive operators as section \ref{hrl} described, $\mathcal{O}_i$ is defined further as $(pre_{\mathcal{O}_i}, eff_{\mathcal{O}_i})$, $pre_{\mathcal{O}_i}$ must be \textit{True} so that operator $\mathcal{O}_i$ is applicable, and $eff_{\mathcal{O}_i}$ is the set of corresponding effects that describe a new environment state after an action $a$ is executed. The state variables we use are listed in Table \ref{table:sv}. Note that for the state variable (sv) \textit{onTop}, the instance \verb|thing| could have the values of both block or slot of the tray (details in Fig. \ref{fig:env}) 


\newcommand{\smalltexttt}[1]{{\scriptsize\texttt{#1}}}

\begin{table}[thpb]
\centering
\caption{Definitions of the state variables and their grounding respectively.}
\scalebox{0.9}{
\begin{tabular}{l|l|l}
\toprule[1pt]
\textbf{State variable} & \textbf{Instances} & \textbf{Grounding} \\
\hline
openGripper  & \smalltexttt{gripper} & gripper is open \\
\hline
graspable & \smalltexttt{gripper, block}  & \makecell[l]{distance between the gripper \\ and block $< 0.01m$ \& \\block between gripper's two fingers} \\
\hline
onTable & \smalltexttt{block, table} & block and table contacts \\
\hline
lifted & \smalltexttt{block} & block height $> 0.01m$ \\
\hline
moving & \smalltexttt{block} & \makecell[l]{block velocity $>0.05m/s$ \& \\block acceleration $< 5m/s^2$}  \\
\hline
above & \smalltexttt{block, thing} & \makecell[l]{block above base block or slot, \\aligned and height difference $>0.05m$} \\
\hline
onTop & \smalltexttt{block1, block2} & \makecell[l]{block1 is on top of block2 \\block2 contacts table}\\
\hline
inSlot & \smalltexttt{block, slot} & \makecell[l]{block is in slot\\block and gripper doesn't contact}\\
\bottomrule[1pt]
\end{tabular}}
\label{table:sv}
\end{table}

\subsubsection{Details about operators} Table \ref{table:pre} shows the relationship between the sv and the operators. Table \ref{table:pre} only shows the must-satisfied conditions for each operator and "-" indicates trivial sv value due to contradiction or dependency between state variables when among in one specific operator, i.e., sv \textit{onTable} in operator \textbf{lift} contradict with sv \textit{lifted}.
\begin{table}[htpb]
\caption{Truth table of the operators' precondition and effect. The blue letter represents precondition and the orange represents effect. A state variable represented with a red-colored letter follows an 'or' logic. In contrast, the relationship between other state variables within the same column is by default considered as 'and' logic.}
\centering
\scalebox{0.8}{
\begin{tabular}[t]{c|c|c|c|c|c|c|c|c|c|c|c|c|c|c}
\toprule[1pt]
\diagbox[width=1.9cm, height=0.7cm]{\textbf{\makecell{sv}}}{\textbf{Operator}} & \multicolumn{2}{c|}{\textbf{open}} &\multicolumn{2}{c|}{\textbf{close}}&\multicolumn{2}{c|}{\textbf{reach}}&\multicolumn{2}{c|}{\textbf{lift}}&\multicolumn{2}{c|}{\textbf{move}}&\multicolumn{2}{c|}{\textbf{stack}}&\multicolumn{2}{c}{\textbf{insert}}\\ \hline
openGripper  & \textcolor{blue}{F}&\textcolor{orange}{T} &\textcolor{blue}{T}&\textcolor{orange}{F} &\textcolor{blue}{T}&-&\textcolor{blue}{F}&\textcolor{orange}{F} & \textcolor{blue}{F}&\textcolor{orange}{F}& \textcolor{blue}{F}&\textcolor{orange}{T}&\textcolor{blue}{F}&\textcolor{orange}{T} \\
graspable & \textcolor{blue}{F}&- & \textcolor{blue}{T}&-& \textcolor{blue}{F}&\textcolor{orange}{T}&\textcolor{blue}{T}&- &\textcolor{blue}{T}&- &\textcolor{blue}{T}&- &\textcolor{blue}{T}&- \\
onTable& \textcolor{blue}{T}&- & \textcolor{blue}{T}&-&\textcolor{blue}{T}&\textcolor{orange}{T}&\textcolor{blue}{T}&\textcolor{orange}{F} &\textcolor{blue}{F}&- & -&\textcolor{orange}{F}&-&\textcolor{orange}{F}\\
lifted &-&- &-&- &-&- & -&\textcolor{orange}{T}&\textcolor{blue}{T}&\textcolor{orange}{T}& \textcolor{blue}{T}&-&\textcolor{blue}{T}&-\\
moving &-&- &-&- &-&- &-&- &-&\textcolor{orange}{T}& \textcolor{blue}{T}&-&-&-\\
above &\textcolor{red}{T}&- &-&- &-&- &-&-&\textcolor{blue}{F}&- &\textcolor{red}{T}&- &\textcolor{red}{T}&-\\
onTop &-&- &-&- &-&- &-&- &-&- &-&\textcolor{orange}{T} &-&\-\\
inSlot &-&- &-&- &-&- &-&- &-&- &-&-&\textcolor{blue}{F}&\textcolor{orange}{T}\\
\bottomrule[1pt]
\end{tabular}}
\label{table:pre}
\vspace{-0.5cm}
\end{table}
\subsubsection{From high-level operators to low-level policies}
Upon the selection of an operator, contingent on the current environment state satisfying one or more preconditions \{$pre_{\mathcal{O}_1}, pre_{\mathcal{O}_2}, \cdots, pre_{\mathcal{O}_K}$ \} set to \textit{True}, the corresponding low-level policies $a$ required for the realization of the chosen operator are learned and optimized by low-level HRL agent as section \ref{hrl} shows. This mapping is from high-level to low-level via a 4-dimensional translation action vector with 3-dim describing the current manipulator's end-effect target position by Cartesian coordinates in the environment, 1-dim used to control the movement of the gripper. 
\subsubsection{Learning method and evaluation}
The proposed learning algorithm is shown in Algorithm \ref{al1}. Here we must point out that $t_{traj}$ has two meanings: one indicates the max total trajectory time steps to terminate the RL episode if the goal is not achieved, and the other means time steps used to successfully achieve the goal within the max total steps. In Algorithm \ref{al1}, \texttt{GetTargets} will find potential targets based on the contact with the table and end effect. \texttt{SolveOpeWithMulObjs} will select a target based on rules: if the current timestep, end-effect contacts one object, then the target will be this object, otherwise, the target will selected randomly from objects that contact with the table. Our method is adapted from RL sandbox\footnote{B. Chan, “Rl sandbox,” https://github.com/chanb/rl\_sandbox\_public, 2020.}, a framework for RL algorithms. Optimization in section \ref{hrl} is done through reparameterization \cite{kingma2013auto}. To mitigate overestimation, clipped double Q-learning \cite{fujimoto2018addressing} is used. Temperature $\alpha$ is also learned for each operator separately. 

When evaluating a single operator's performance, things are different from the training process since there's no planning now, which means we don't know which target is the agent's choice, especially when multiple targets are involved (i.e., operator stack is between 2 blocks), result in obstacles when computing reward since the reward function need to know the specific targets. So in evaluation, first, whether the current operator is executed successfully will be judged. This will be applied to all objects in the environment. If the operator succeeds on one object or two objects, the reward will be calculated for certain object/objects, otherwise will be calculated for all objects, and the maximum reward vector as the current time steps' reward value. Other details are shown in Algorithm \ref{al2}.
\subsubsection{Reward and success criteria}
This section introduces the reward function and success criteria used in our method. Here $a_g$ refers to gripper action (1-dim), and $a$ is the translational position shifts of end-effect (3-dim). Generally, reward functions contain both sparse rewards and dense rewards because dense rewards could boost learning speed \cite{ablett2023learning}. The following elements are used to formulate different operators' reward functions: 1) binary check for gripper action; 2) shaping term on magnitudes of $a$; 3) sparse reward based on the distance between end-effector and a block; 4) dense reward-based on the distance between the end-effector and a block; 5) binary check for the distance between the gripper's two fingers; 6) dense block height reward; 7) sparse block height reward; 8) block velocity reward; 9) block acceleration penalty; 10) dense reward based on the distance between the target position and the block's current position. Note, that the sparse reward generally is binary. The detailed reward function structures are listed in Table \ref{table:reward}.

\begin{table}[htp]
\centering
\caption{Reward and success criteria for each operator.}
\scalebox{0.85}{
\begin{tabular}[t]{c|l|l}
\toprule[1pt]
 & \makecell[l]{\textbf{Reward function}\\ \textbf{elements}} & \textbf{Success criteria} \\ \hline
\textbf{open} &1, 2, 3 & $a_g<0$\\ \hline
\textbf{close} &1, 2, 3  & $a_g>0$\\ \hline
\textbf{reach} &4 &distance between block and end-effect $<0.02m$\\ \hline
\textbf{lift} & 1, 4, 5, 6& block height $>0.04m$\\ \hline
\textbf{move} &3, 4, 5, 7, 8, 9&block vel. $>0.05m/s$ acc. $<5m/s^2$\\ \hline
\textbf{stack} &1, 2, 3, 4, 5, 7, 8, 10&\makecell[l]{1. all blocks contact\\2. height between 2 contact blocks $>0.035m$\\3. only one block contact with tray}\\ \hline
\textbf{insert} &3, 4, 10 &distance between block and slot $<0.003m$\\
\bottomrule[1pt]
\end{tabular}}
\label{table:reward}
\end{table}

\vspace{-0.5cm}
\IncMargin{1em}
\begin{algorithm}
\footnotesize
\SetKwInOut{Input}{input}\SetKwInOut{Output}{output}
  \Input{Planner period $\xi$, sample batch size $N$, trajectory time steps $t_{traj}$}
  \BlankLine
  \emph{Initialize reply buffer $\mathcal{B}$}\;
  \While{$t<t_{traj}$}{
      \emph{Interact with environment}\\
      \emph{For every $\xi$ steps}\\
      \eIf{$\pi_{\mathcal O_{t-1}}$ achieved}{
      $pre_{\mathcal O_{t}}\leftarrow eff_{\mathcal O_{t}}$\\}
      {
      envState $\leftarrow$\texttt{ReadEnv}\\
      targets $\leftarrow$\texttt{GetTargets}(envState)\\
      $pre_{\mathcal O_{t}}\leftarrow$\texttt{ComputeCondition}(targets)
      }
      $\pi_{\mathcal O_{t}}\leftarrow$\texttt{GetOperator}($pre_{\mathcal O_{t}}$)\\
      
      $a_t\sim\pi(a_0|s_t,\mathcal{O}_{t})$\\
      \emph{Execute $a_t$ and observe next envState $s'_t$}\\
      \emph{Store transition $\left \langle s_t,a_t,s'_t \right \rangle$ in $\mathcal{B}$}\\
      Sample $\{(s_i,a_i)\}_{i=1}^N\sim \mathcal{B}$\\
      \emph{Judge whether $\mathcal{O}_t$ is successfully executed}\\
      \For{$\mathcal{O}\leftarrow 0$ \KwTo $K$}{
      \emph{Compute reward $\vec{r}(s_t,a_t)$}
      }
      \emph{Update $\pi$ and $Q$ following Eq. (5) and Eq. (6)} 
      $t=t+1$
      }
  \caption{Training and Learning}\label{al1}
\end{algorithm}\DecMargin{1em}
\IncMargin{1em}
\begin{algorithm}
    \footnotesize
  \SetKwInOut{Input}{input}\SetKwInOut{Output}{output}
  \Input{Planner period $\xi$, sample batch size $N$, max. evaluation time steps for all operators $t_{tot}$, single object operator $\mathcal O_{single}$, multi objects operator $\mathcal{O}_{mul}$}
  \BlankLine
  \emph{for each operator $\mathcal{O}_1,\cdots,\mathcal{O}_i,\cdots,\mathcal{O}_K$, subscript $i$ is operator id}\\
  \While{$t<t_{tot}$}{
      \emph{Interact with environment}\\
      $i \leftarrow$ \texttt{findFirstIndexWhere}($count < t_{tot}$)\\
      $count \leftarrow count + 1$\\
      \emph{match $i$ with operator, get $\mathcal{O}_i$}\\
      \eIf{$\mathcal{O}_i$ in $\mathcal{O}_{single}$}{
      $\pi_{\mathcal{O}_i}\leftarrow \mathcal{O}_i$
      \\}{
      $\mathcal{O}_{i-1}$, operatorState $\leftarrow$operatorStateDict\\
      envState $\leftarrow$ \texttt{ReadEnv}\\
      targets $\leftarrow$ \texttt{GetTargets}(envState)\\
      $\pi_{\mathcal{O}_i}\leftarrow$ \texttt{SolveOpeWithMulObjs}(targets, $\mathcal{O}_{i-1}$)\\
      }
      $a_t\sim \pi(a_0|s_t,\mathcal{O}_{i})$\\
      \emph{Execute $a_t$ and observe next envState $s'_t$}\\
      \emph{Judge whether $\mathcal{O}_i$ is successfully executed}\\
      \For{$\mathcal{O}\leftarrow 0$ \KwTo $K$}{
      \emph{Compute reward $\vec{r}(s_t,a_t, \mathcal{O}_i)$}
      }
  }
  \caption{Evaluation for single operators}\label{al2}
\end{algorithm}\DecMargin{1em}
\vspace{-0.5cm}
\section{Experiment}\label{exp}
\begin{figure}[htp]
  \centering
  \includegraphics[width=0.45\textwidth]{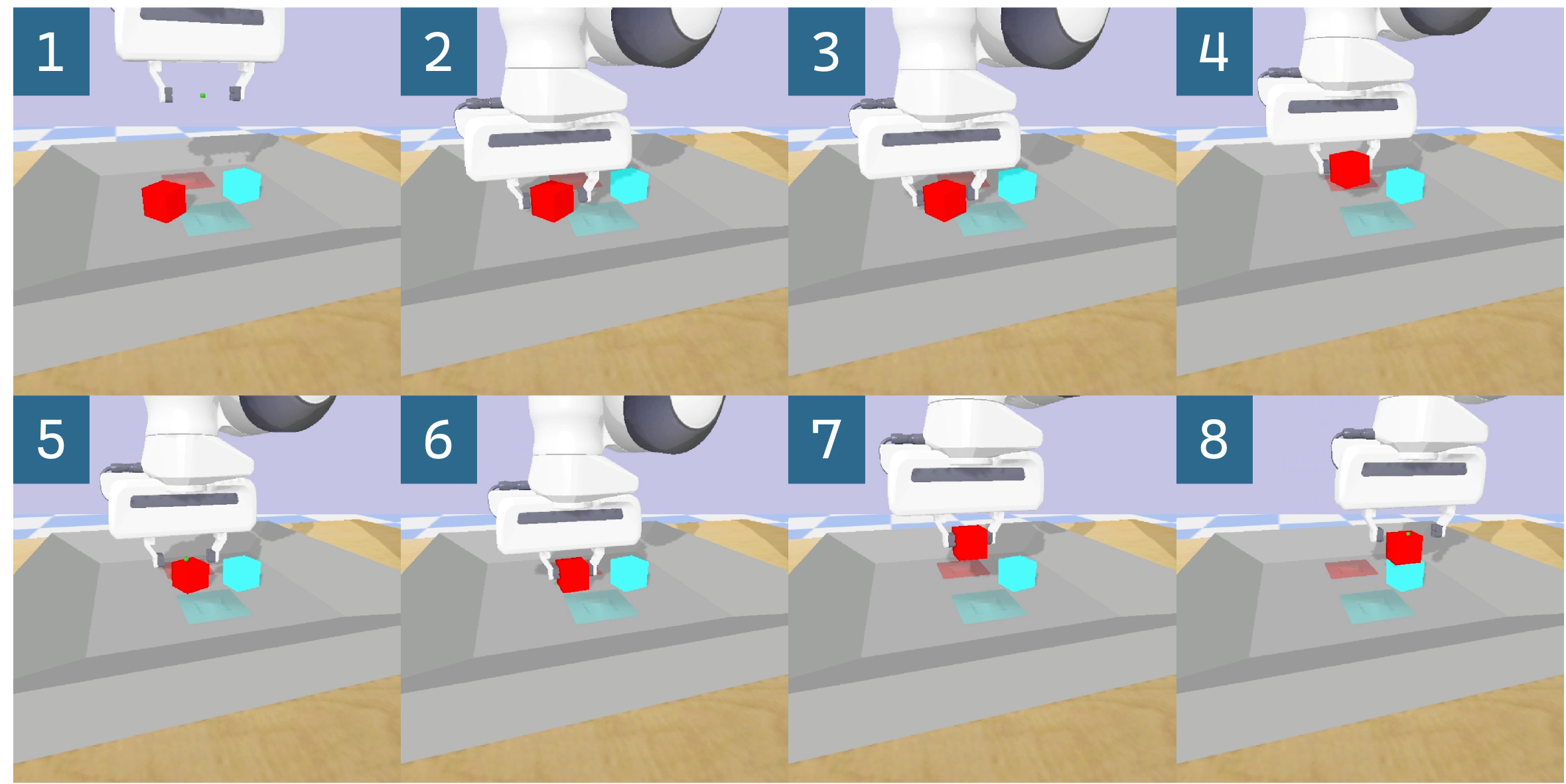}
  \caption{Example of executing the learned sequence and policies for the task of stacking 2 blocks. At step $1$, the agent \textbf{reaches} the red block, then at step $2$, since \textbf{reach} is finished, the next operator would be \textbf{close} (step $3$), then \textbf{lift} is executed (step $4$), but at step $5$, the unexpected situation happens that the block is dropped, so it needs to re-\textbf{reach} ($6$) and \textbf{lift} it ($7$), finally the agent successfully \textbf{move} the block and \textbf{stack} it on the blue one ($8$), now the whole task of STACK is finished. \href{https://youtu.be/FQYWLhLTeOg}{\texttt{https://youtu.be/FQYWLhLTeOg}}.} 
  \label{fig:env}
\end{figure}
\subsection{Environment setting}
The environment was adapted from the manipulator-learning \footnote{T. Ablett, “manipulator-learning,” https://github.com/utiasSTARS/
manipulator-learning, 2022.} (see Fig. \ref{fig:env}), with Franka Emika Panda manipulator, two blocks for each has the shape of $4 cm\times4 cm\times 4 cm$, one tray ($30 cm \times 30 cm$) with two slots ($4.1 cm\times4.1 cm\times1 cm$) and sloped edges that keeps blocks within reachable workspace, a table. The end-effector's range of motion, measured from the tool center point situated directly between the gripper fingers, is confined within a box measuring $30 cm \times 30 cm \times 30 cm$. The lower limit of this boundary is set to enable the gripper to engage with objects while avoiding collision with the tray's bottom. 

Actions related to environmental movement involve three degrees of freedom (3-DOF) for translational shifts in position. These shifts are calculated relative to the present location of the robotic arm's end-effector. To achieve this, PyBullet's integrated function for position-based inverse kinematics is used, which helps us generate the appropriate joint commands for the robotic arm.
Besides these three dimensions, action space includes a fourth dimension dedicated to the gripper's operation. This extra dimension is designed to be compatible with policy models that produce solely continuous output values. Specifically, any real number greater than zero instructs the gripper to open, while a number less than zero commands it to close.

The system operates at a rate of $20 Hz$, for stack $2$ blocks, each training episode is capped at $18$ seconds, which equates to $360$ time steps for each episode. The positions of the blocks within the tray are randomized between episodes. Additionally, the end-effector's initial position is randomized to lie anywhere between $5 cm$ and $14.5 cm$ above the tray, adhering to previously defined bounds. Moreover, the gripper starts in a fully open state at the beginning of each episode.

The operators we used in experiments are listed in Table \ref{table:pre}: open(gripper), close(gripper), reach, lift, move, stack and insert with their own preconditions and effects. The capital letters indicate the name of the whole planning process, for example, "STACK" refers to stacking 2 blocks, which consists of many operators (e.g. lowercase reach, stack, etc.), i.e., STACK = \{reach, move, ..., stack\}. This symbolic structure of operators is a set of causal rules that outline the features of objects and environment, given by human experts and hard-coded. 

During the training process, after the execution of one operator, there would be two possible situations: success or failure. If successful, the next operator will be generated otherwise the current operator would execute for at most 45 time steps.
\subsection{Experiment results}
We assess the efficacy of our proposed approach through a dual-metric evaluation framework, focusing on both the chained policy execution success rate and the individual operator success rate \footnote{Code is available at: \href{https://gitlab.com/craft_lab/rl-and-symbolic_planning/RL_operator}{\texttt{https://gitlab.com/craft\_lab/rl-\\and-symbolic\_planning/RL\_operator}}}.
\subsubsection{Chained policy execution success rate}

\begin{figure}[ht]
\begin{minipage}[t]{0.48\textwidth}
\centering
\includegraphics[width=0.48\textwidth]{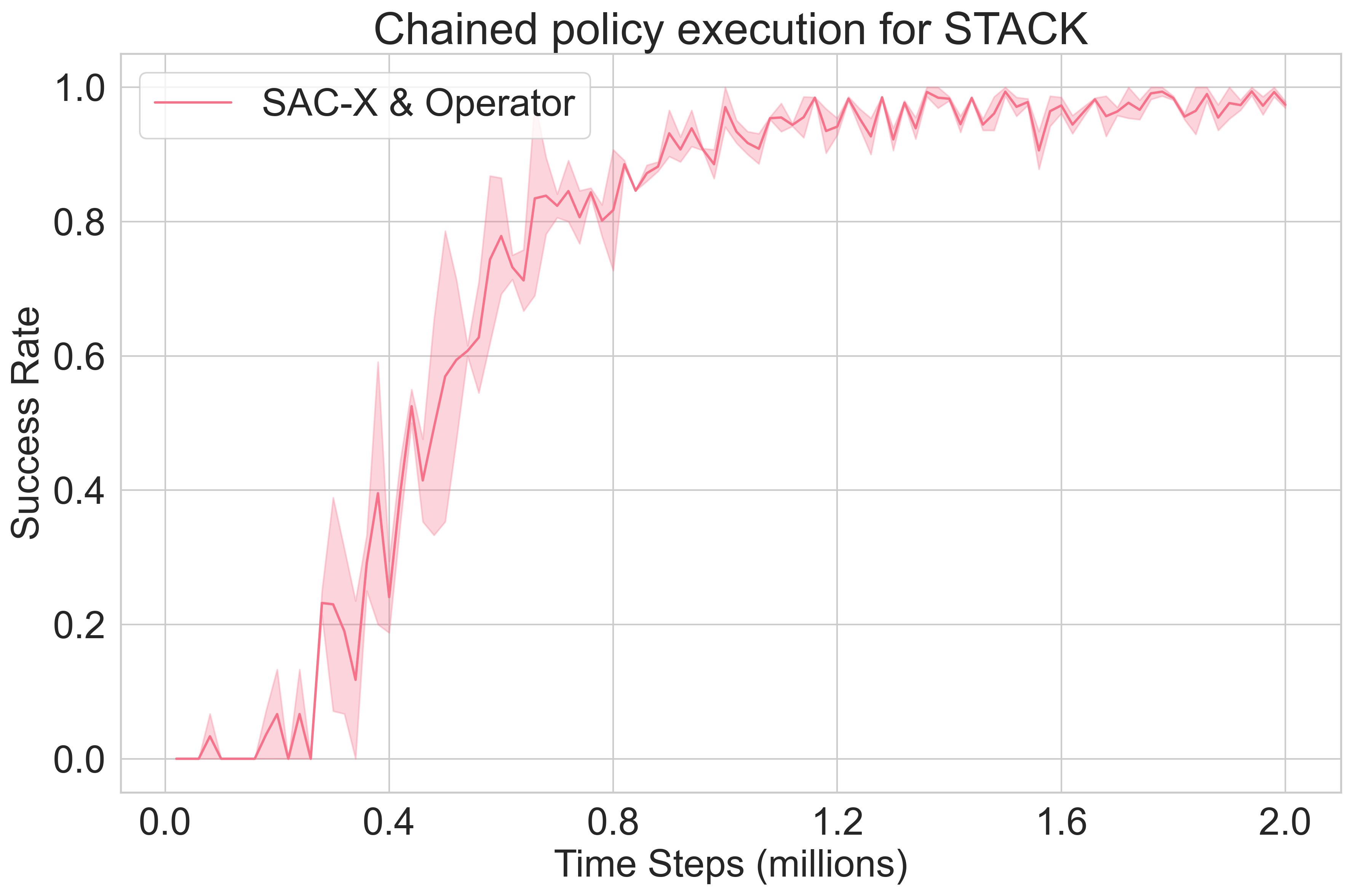}
\includegraphics[width=0.48\textwidth]{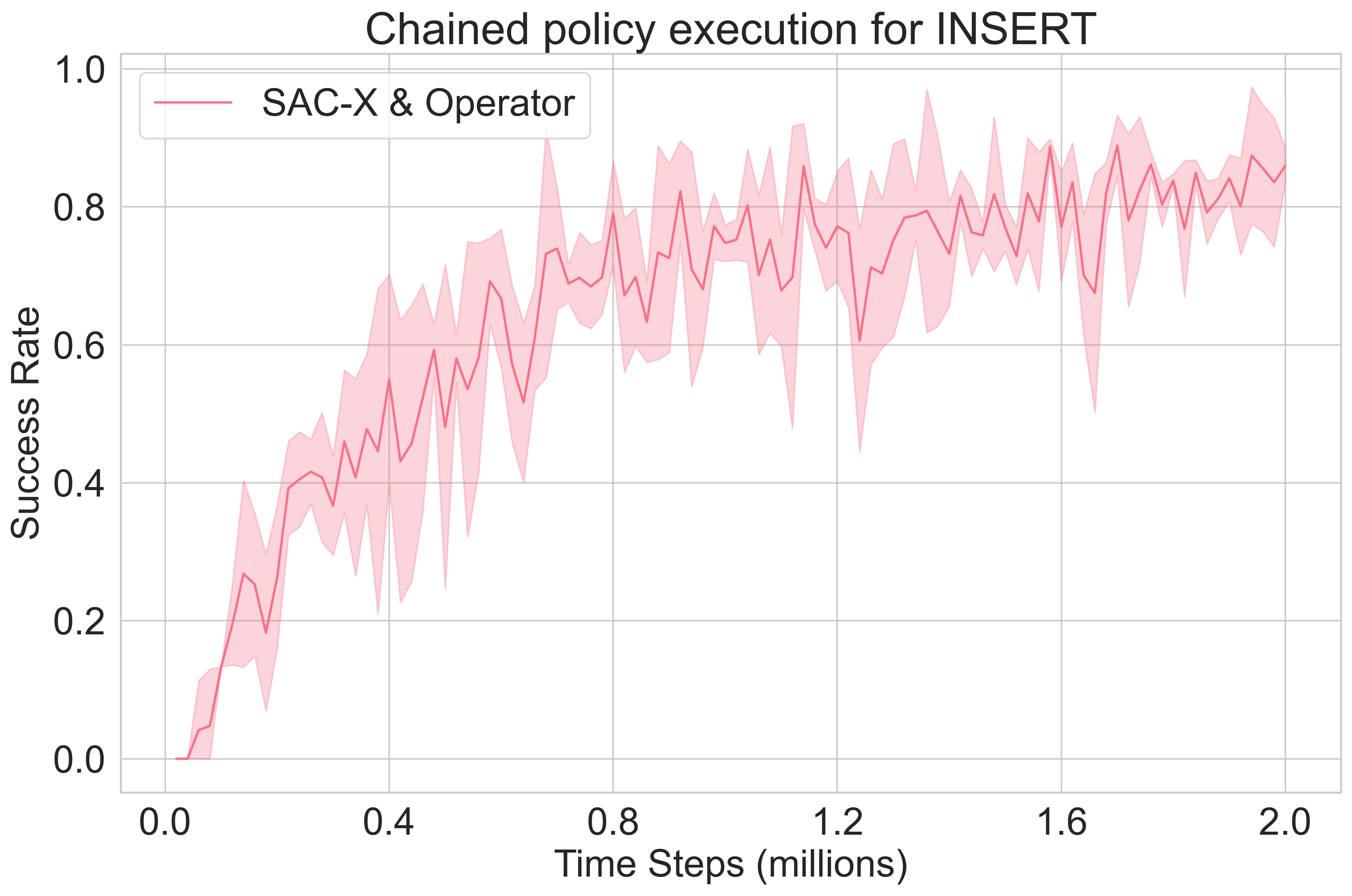}
\caption{Chained policy execution success rate for STACK 2 blocks and INSERT block into a slot.}
\label{fig:planning}
\end{minipage}
\end{figure}
Chained policy execution for STACK and INSERT involves a sequence of operators. The construction of STACK is based on a specific set of operators: $\{open, close, reach, lift, move, stack\}$. We reuse the high-level structure to organize the order of operators for various tasks. For instance, in the task of INSERT, we introduce an additional special operator, $insert$, to create holistic plans based on preconditions and effects, as presented in Table.\ref{table:pre}. Thus, INSERT is constructed using the set of operators $\{open, close, reach, lift, move, insert\}$. By introducing task-specific operators and simple logical judgments in the high-level structure, it is possible to efficiently scale and adapt to new tasks without having to redesign the entire system.

We test our method on 3 different seeds, and the chained policy execution success rate is presented in Fig.\ref{fig:planning}. The experiment result achieved a near-perfect success rate of $97.2\%$ for chained policy to stack two blocks after training for $1.5M$ environment interaction steps. Fig.\ref{fig:env} represents a typical example that shows how our method use chained policy for STACK. To ensure the planning process is stable, only if one operator succeeds at the current frame and lasts for some time, we define this execution as success. The holding time restriction for all operators is listed in Table \ref{table:restricts}. 

INSERT is a task that requires a manipulator to insert one block into the slot, the slot area is shown in the right of Fig.\ref{fig:env} which is light-colored on the tray. The chained policy execution success rate would reach up to around $85\%$.

\vspace{-2mm}
\begin{table}[thb]
\centering
\caption{Holding time restriction for each operator.}
\scalebox{0.9}{
\begin{tabular}[t]{cccccccc}
\toprule[1pt]
{\textbf{Operator}} & \textbf{open} & \textbf{close} &\textbf{reach} &\textbf{lift} &\textbf{move} &\textbf{stack} &\textbf{insert}\\ \hline
\textbf{Time restrict/$s$}   & 0.2 & 0.2 & 0.1 & 0.3 & 0.2 & 0.5 & 0.5  \\
\bottomrule[1pt]
\end{tabular}}
\label{table:restricts}
\end{table}\
\subsubsection{Individual operator success rate}
\vspace{-2mm}
\begin{figure}[ht]
\begin{minipage}[t]{0.48\textwidth}
\centering
\includegraphics[width=0.32\textwidth]{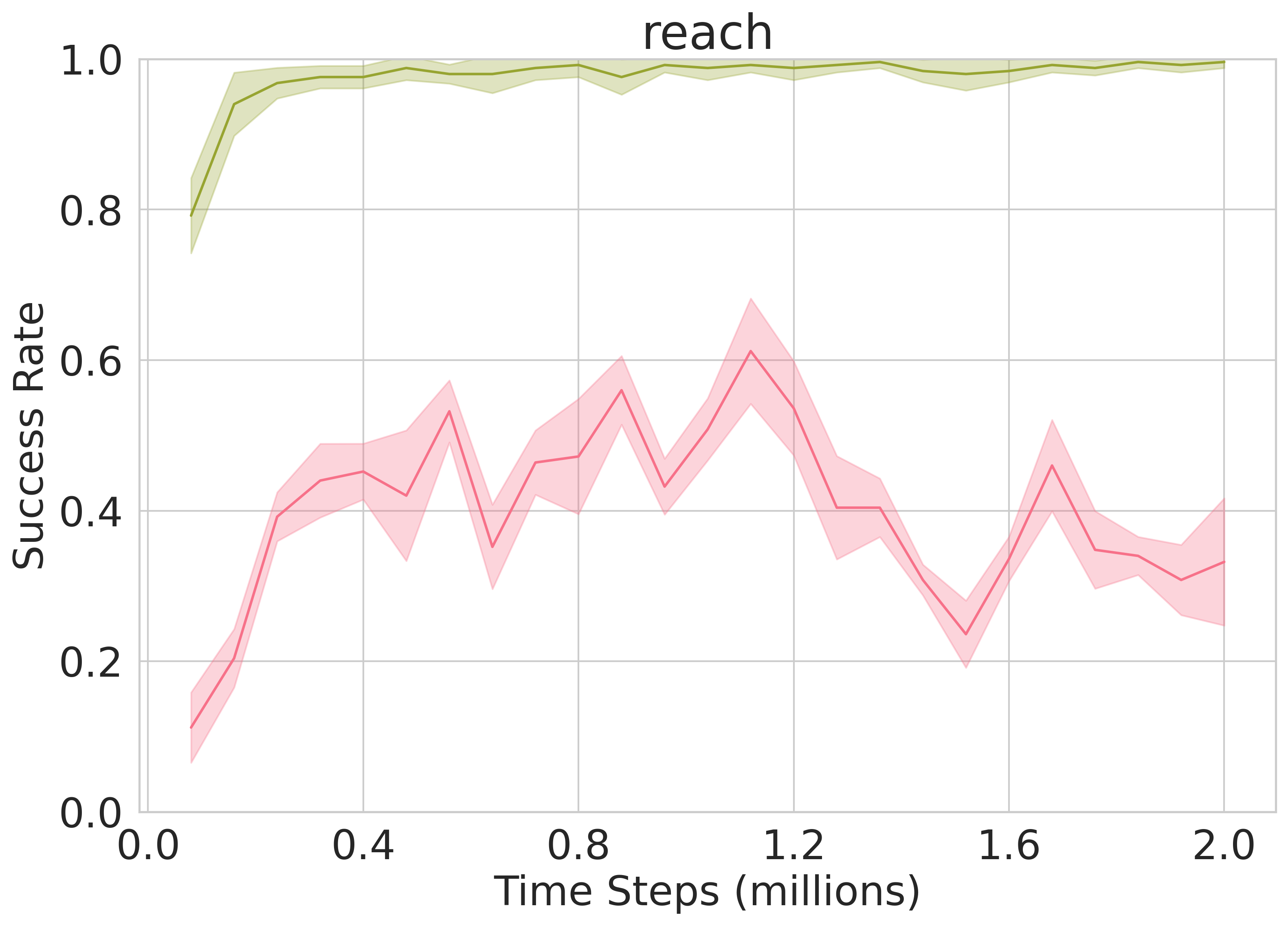}
\includegraphics[width=0.32\textwidth]{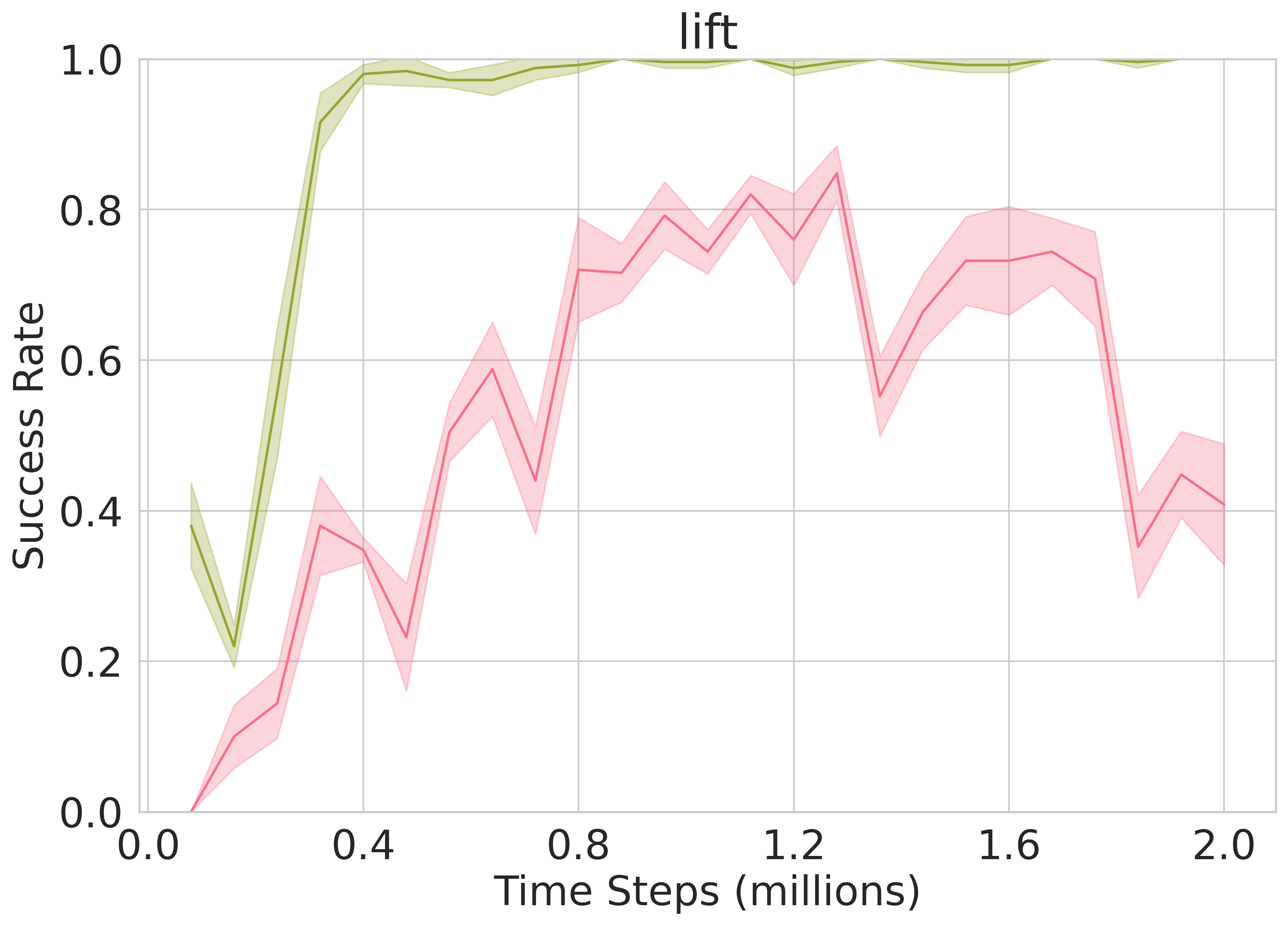}
\includegraphics[width=0.32\textwidth]{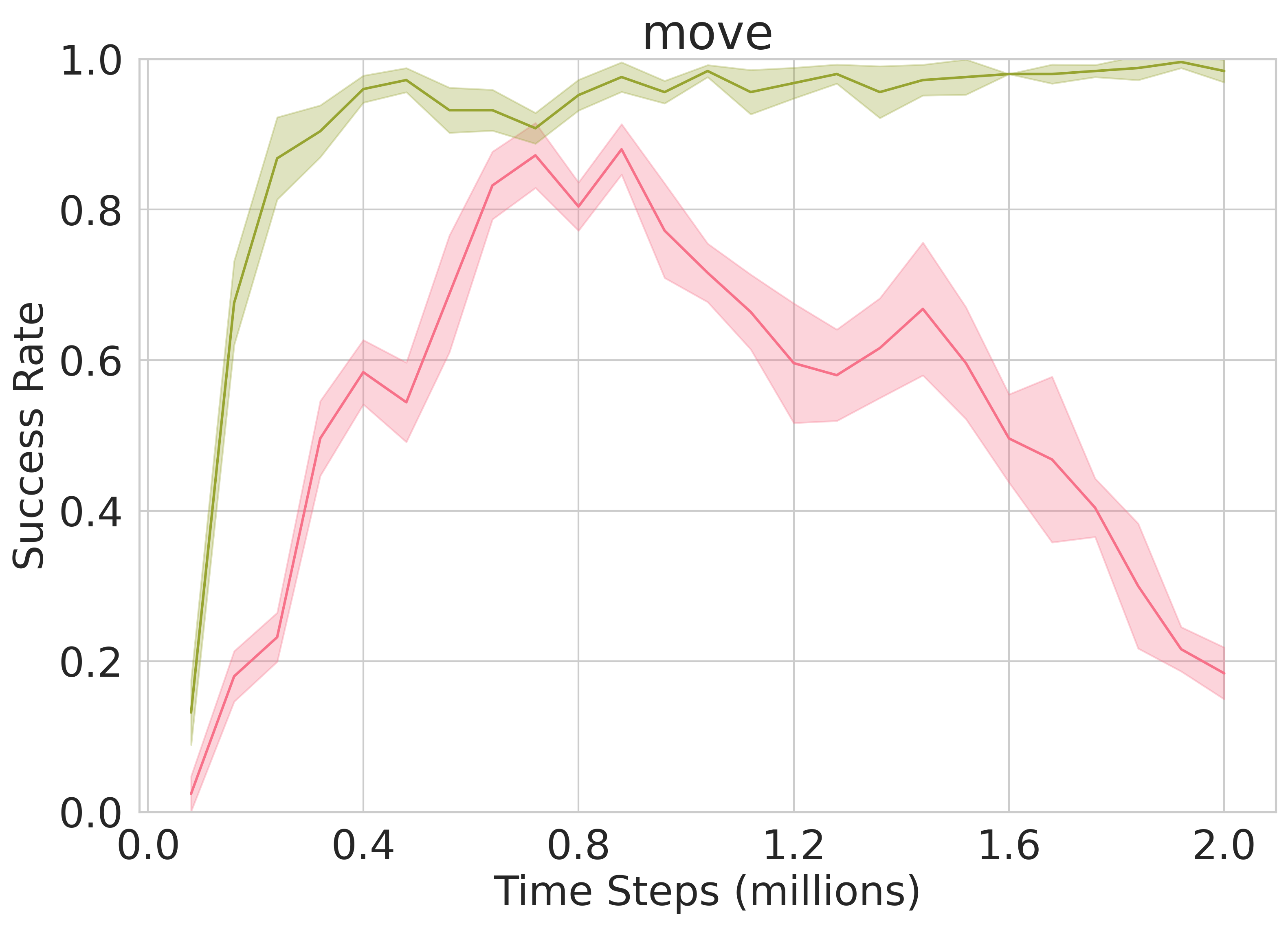}
\caption{Single operator success rate for easy tasks. Fig. \ref{fig:easy} and Fig. \ref{fig:hard} share the same legend notation.}
\label{fig:easy}
\end{minipage}
\begin{minipage}[t]{0.48\textwidth}
\centering
\includegraphics[width=0.32\textwidth]{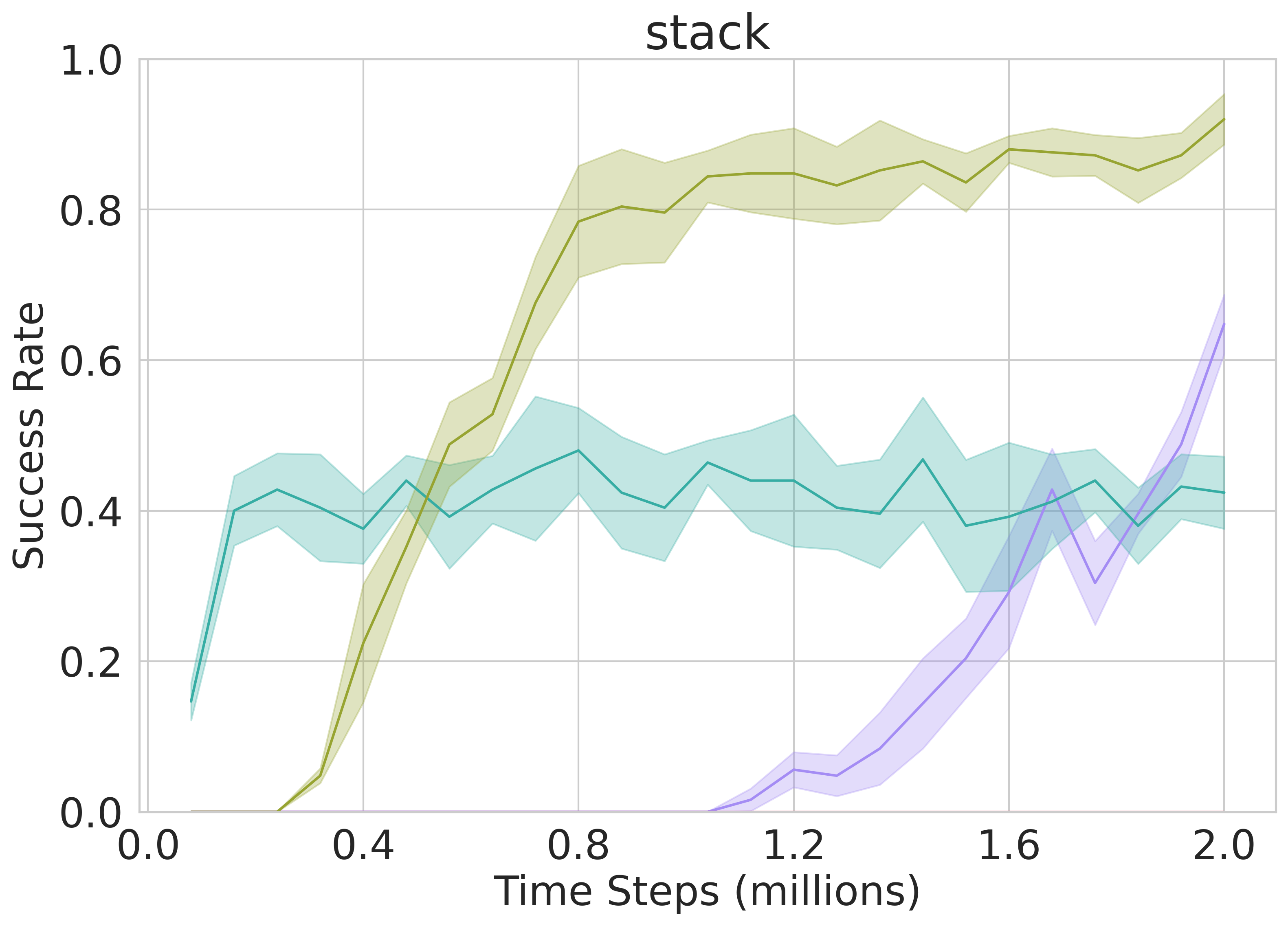}
\includegraphics[width=0.32\textwidth]{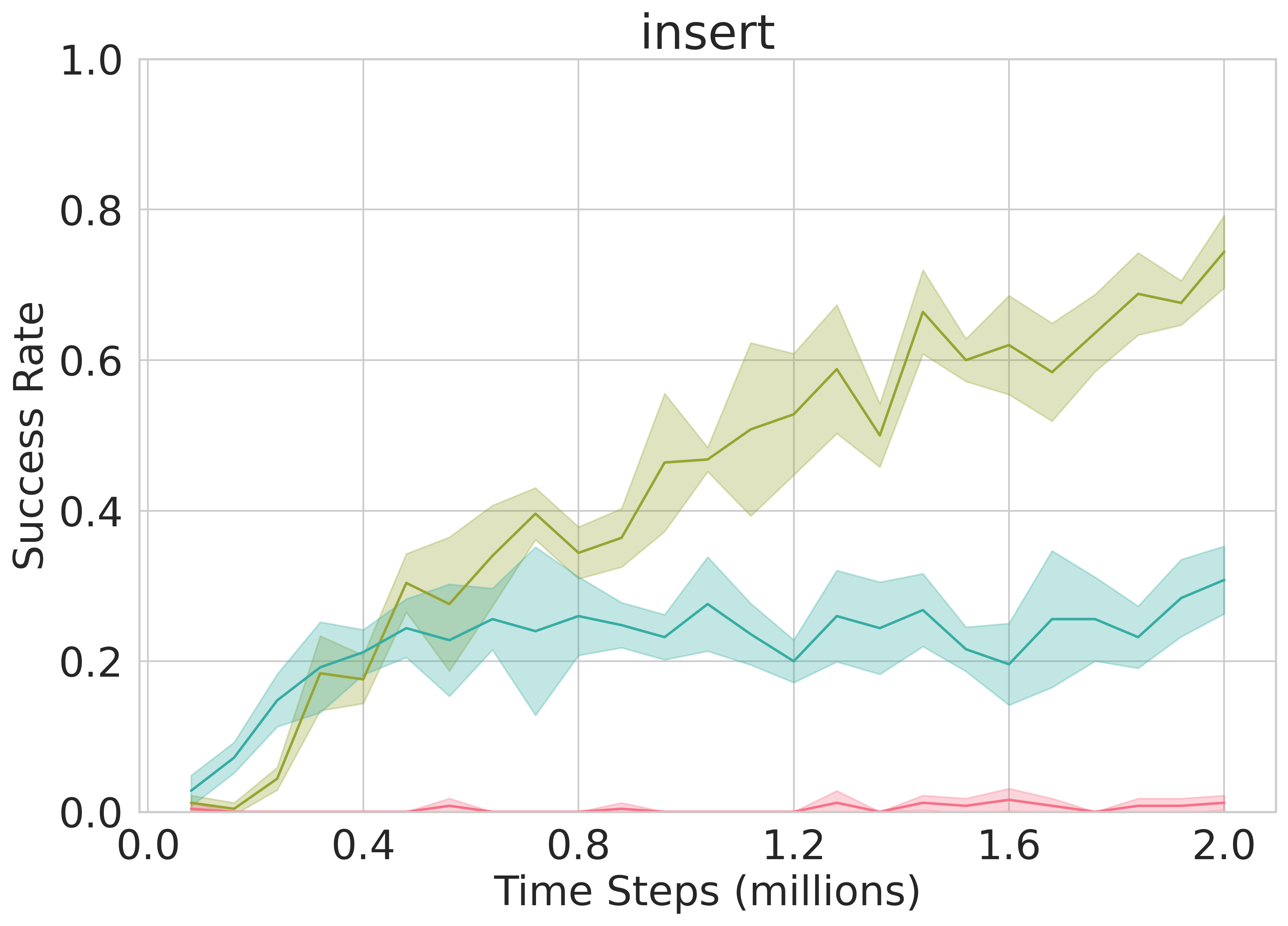}
\raisebox{0.5cm}{\includegraphics[width=0.2\textwidth]{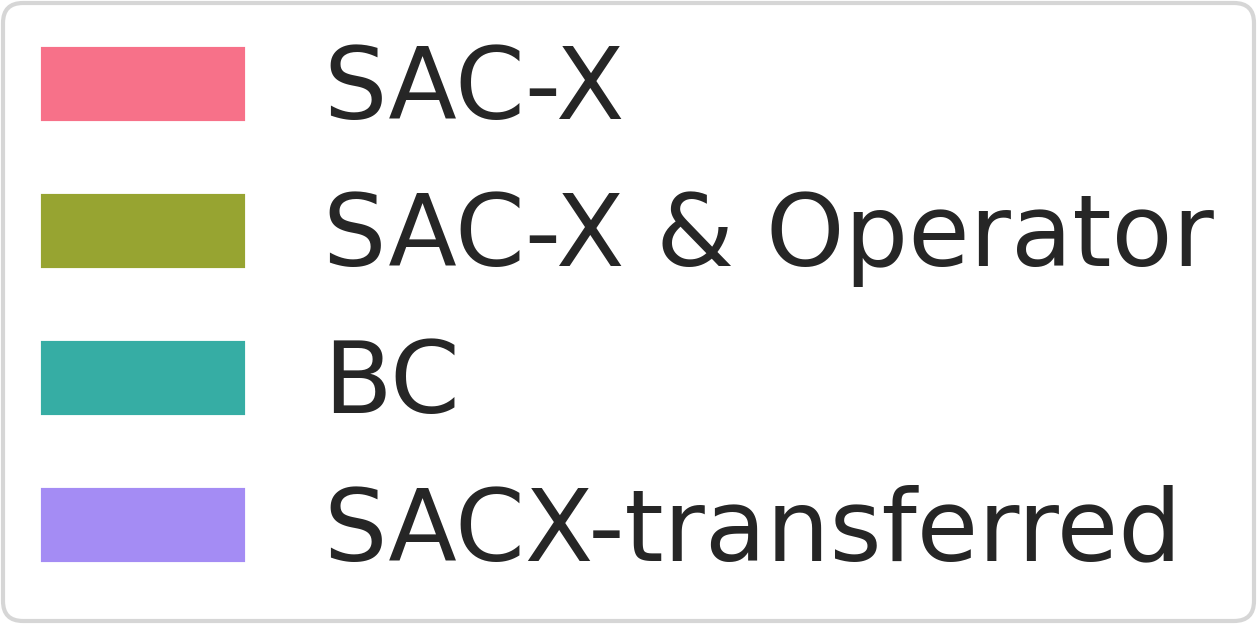}}
\caption{Single operator success rate for long-horizon tasks. In the stack figure, since training SAC-X from scratch could not learn to manage this task, as figures showed, the success rate is always 0, we then pre-trained the SAC-X model to learn "move" for $1M$ time steps, then used this model to apply transfer learning to learn "stack". Similarly for the task "insert".}
\label{fig:hard}
\end{minipage}
\end{figure}
Since our method also learns low-level policy, thus makes it possible to evaluate individual operators, which also can be treated as a single policy. In Fig. \ref{fig:easy} and \ref{fig:hard}, performance comparisons between our method and other algorithms are shown. 

For single low-level control policies like reach, lift and move in Fig. \ref{fig:easy}, both our method (SAC-X \& Operator) and SAC-X \cite{riedmiller2018learning}\footnote{We used the adapted version of \cite{ablett2023learning} which made it compatible to learn the hierarchical agent.\label{sacx}} learned the policies successfully, but our method is superior on training time and success rate, after training for $0.8M$ time steps, the success rate on average is: reach - $98.9\%$, lift - $99.7\%$, move - $97.4\%$. Our method shows a high task performance within a very short period and also a much more stable performance.

As for long-horizon manipulation as stack and insert, original SAC-X \cite{riedmiller2018learning} needs $64.8M$ to learn (thus not included here), SAC-X \footref{sacx} (with a heuristic planner) didn't manage to learn the tasks (see Fig. \ref{fig:hard}), which further demonstrates our hypothesis discussed in section \ref{combine}. The heuristic planner employs conditional probabilities based on their current context and historical performance to guide the selection and sequencing of tasks. We also specially trained a transferred version of SAC-X, pre-train a feasible "move" model for $1M$ time steps, then used this model to train "stack" like \cite{ablett2023learning} did, they also found the plateau problem. 

Though our method didn't use human demonstrations directly, symbolic representation and relations are knowledge that comes from humans, thus drawing an implicit relation with human experts, so we also compare our method with imitation learning baseline BC, which learned from human demonstrations (gathered from an RL expert \cite{ablett2023learning}) explicitly. 

Fig. \ref{fig:env} (fig. label $8$) depicts the outcome of the operator stack in our method. The results showed in Fig. \ref{fig:hard} that our method's success rate for stack quickly surpassed BC and SAC-X transferred whilst maintaining a high rate, with the highest at $92\%$ and an average of $85\%$ after training for $0.8 M$ time steps. Note it turns out that even through transfer learning, SAC-X\footref{sacx} still needs $2.5M$ (including pre-train) to learn stack to achieve a success rate over $80\%$. An additional experiment on the insert task also illustrates a better performance, and the success rate for our method would even increase if trained for a longer time. All experiments are tested over 5 different seeds.


\vspace{-0.51mm}
\section{Discussion}
The performance comparisons between our method and the heuristic baseline (including SAC-X and SAC-X transferred), shown in Figs.\ref{fig:easy} and \ref{fig:hard}, further highlight the limitations of the high-level heuristic planner in adaptability to dynamic environments. One plausible reason for this shortfall is the over-reliance on predefined task dependencies, which may overlook the potential for exploring new strategies and potentially offering suboptimal guidance to low-level policies due to their static nature. In contrast, our method dynamically selects the most appropriate policy by integrating the latest environment information and leveraging real-time feedback, thereby enabling more informed and adaptive task planning. Furthermore, despite SAC-X and SAC-X transferred showing some capacity to leverage prior knowledge, they fall short when dealing with complex manipulation tasks, while our method, even when the manual definition of the operators (see Table \ref{table:pre}) is needed, improves the success rate in such scenarios, highlighting the critical importance of adaptive sequencing of low-level policies in dynamic and unpredictable environments.
\vspace{-0.5mm}
\section{Conclusion}
Our method tackles the challenge of long-horizon tasks in robotic manipulation by integrating symbolic planning with Reinforcement Learning methods through operators. We introduce dual-purpose high-level operators, contributing both to generate holistic sequencing of actions with the possibility of executing them as independent policies. Experiments confirm a $97.2\%$ STACK chained policy execution success rate and $80\%$ for INSERT, also even an $85\%$ average success rate for individual operators in the absence of a holistic planning context. Our method thereby offers a promising avenue for achieving both high task performance and efficient learning in the complex domain of robotic manipulation.


\addtolength{\textheight}{-12cm}   





\bibliographystyle{IEEEtran}
\bibliography{ref}
\end{document}